\documentclass[letterpaper]{article}

\usepackage{spconf,amsmath,graphicx,bm,amsmath,amssymb,mathrsfs}

\title{Conditional MoCoGAN for Zero-Shot Video Generation}
\name{Shun Kimura and Kazuhiko Kawamoto}
\address{Chiba University, Japan}
\begin{document}
\topmargin=0mm
%
\maketitle
\begin{abstract}
We propose a conditional generative adversarial network (GAN) model for zero-shot video generation. 
In this study, we have explored zero-shot conditional generation setting.
In other words, we generate unseen videos from training samples with missing classes.
The task is an extension of conditional data generation.
The key idea is to learn disentangled representations in the latent space of a GAN.
To realize this objective, we base our model on the motion and content decomposed GAN and conditional GAN for image generation.
We build the model to find better-disentangled representations and to generate good-quality videos.
We demonstrate the effectiveness of our proposed model through experiments on the Weizmann action database and the MUG facial expression database.
\end{abstract}

\section{Introduction}
\label{sec:intro}
Video recognition using deep learning has been actively researched and is being used in the real world. 
Human action recognition from videos has been applied in various fields, and further development is expected.
Deep learning requires a rich and accurate dataset.
However, when we try to form a large amount of data, we often encounter the problems of insufficient data and class imbalance.
One method for solving these problems is using deep generative models.
Generative adversarial networks (GANs) \cite{gan} are deep generative models that can generate realistic data based on a learned dataset.
The generation of realistic static images has progressed significantly by using deep generative models such as GANs. 
In recent years, GANs have been applied to video generation.
Typical deep video generation models include VideoGAN \cite{videogan} and the motion and content decomposed GAN (MoCoGAN) \cite{mocogan}.

Conditional generation, which generates data of a specified class, is crucial to solving the problems mentioned above using deep generative models, such as GANs.
Zero-shot generation, which generates the data of a class that has never been seen in training from the information contained in the training data, can be considered as an extension of conditional generation.
For example, if the training data contain only a walking video of person A and a dancing video of person B, this task generates a dancing video of person A.

There have been several studies on zero-shot generative models.
DistillGAN \cite{distillgan} showed that it is possible to generate zero-shot images by providing captions of the classes that are not included in the training data to a model trained on a domain.
Research on zero-shot video generation includes HOIGAN \cite{hoigan} and \textit{Everybody dance now} \cite{ebdn}. 
In the latter, the video is converted from the source video to the target video through frame-by-frame image conversion using the skeletal information of the video containing the person.
HOIGAN achieved zero-shot video generation by conditioning on a combination of actions and objects that were not included in the training data.
However, its architecture is too complex because it uses word embedding vectors and masking images, and it has multiple discriminators with different roles. The same is true for \textit{Everybody dance now} \cite{ebdn}, which requires frame-by-frame skeletal information of the source video and a dedicated image transformation network from the skeleton to the target.
The conditioning for video generation in these methods is very strong.

In this study, we propose a class-conditional video generative model and tackle the task of zero-shot video generation by conditioning on classes that are not included in the training data. To achieve class-conditional video generation, we incorporated the learning method of conditional GAN (CGAN) \cite{cgan}, which is a class-conditional generation model, into the training of MoCoGAN.
To summarize, the contributions of this study are as follows:
\begin{itemize}
\item{The class-conditional generation method for static images is also effective in GANs for video generation.}
\item{When the class of a video is decomposed into two classes, namely, motion and content, it can correctly generate a video that satisfies both classes simultaneously, even if the combination of classes is not included in the training data.}
\end{itemize}

\begin{figure*}[h!]
    \centering
    \centerline{\includegraphics[width=15.4cm]{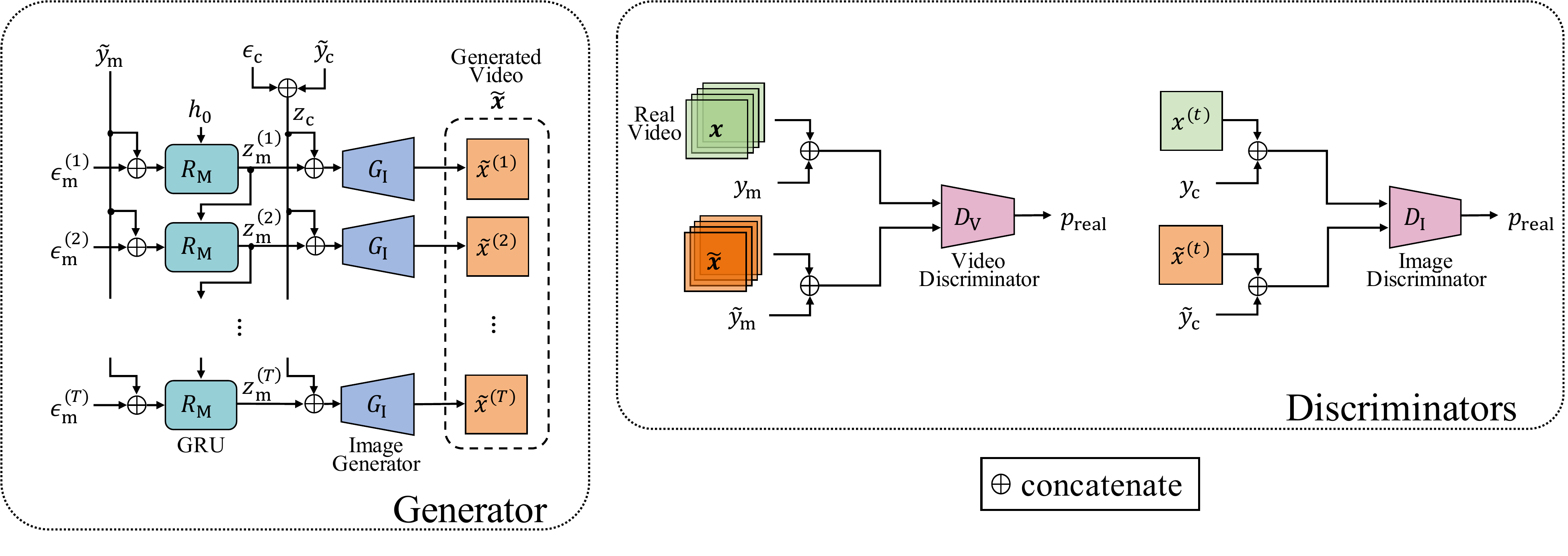}}
  \caption{{\bf Architecture of conditional-MoCoGAN}}
  \label{fig:ourmodel}
\end{figure*}

\section{Related work}
\label{sec:works}
GANs have attracted significant attention as deep generative models, and various derived models have been proposed.

\subsection{Generative Adversarial Networks (GANs)}
The GAN architecture consists of two networks: a generator and a discriminator.
The generator $G$ takes a noise vector $z$ as input and tries to generate data that resemble the training data.
Meanwhile, the discriminator $D$ takes data as input and tries to determine whether they are the real training data, or the fake data generated by $G$.
The GAN trains these two networks via an adversarial process given by the following objective:
\begin{eqnarray}
  \label{form:gan}
  \min_{G} \max_{D}
  \mathbb{E}_{x}[\log{D(x)}]
  +\mathbb{E}_{\tilde{x}}[\log{(1-D(\tilde{x}))}], 
\end{eqnarray}
where ${\tilde x} = G(z)$, and $x$ represent the training data.

\subsection{Conditional GAN (CGAN)}
CGAN is a conditional generative model that is based on the basic GAN structure.
The main difference between the CGAN and the original GAN is that both the noise vector and the condition vector are given as input to the generator and discriminator.
The objective function is expressed as follows:
\begin{eqnarray}
  \min_{G} \max_{D}
  \mathbb{E}_{x}[\log{D(x|y)}] 
  +\mathbb{E}_{\tilde{x}}[\log{(1-D(\tilde{x}|\tilde{y}))}], 
\end{eqnarray}
where $y$ and ${\tilde y}$ are the class labels of $x$ and ${\tilde x}$, respectively.


\subsection{Motion and Content Decomposed GAN (MoCoGAN)}
MoCoGAN is a model that applies a GAN to video generation.
It assumes that a video can be decomposed into two features, namely, motion and content, and it generates a video by sampling each input noise vector from a different latent space.
The generator $G_{\rm I}$ generates a single image $\tilde{x}^{(t)}$ from the given input vector.
This image is one frame of the video, and by inputting $T$ vectors, $\tilde{\bm{x}}$, a $T$-frame video is generated.
The input vector is divided into ${z}_{\rm c}$, which is a vector representing the content of the video, and $[{z}_{\rm m}^{(1)},\cdots,{z}_{\rm m}^{(T)}]$, which are vectors representing the motion of the video.
Because it is more natural to have uniform content within a single video, the content vector is sampled only once per video, and it is combined with motion vectors generated by the recurrent model for each frame.
We input $[[{z}_{\rm c}, {z}_{\rm m}^{(1)}],\cdots,[{z}_{\rm c}, {z}_{\rm m}^{(T)}]]$ to $G_{\rm I}$ and generate frames from each vector.

The whole discriminator consists of two discriminators, $D_{\rm I}$ and $D_{\rm V}$.
$D_{\rm I}$ takes a randomly sampled frame from the video as input, and $D_{\rm V}$ takes the entire video as input.

The objective function is expressed as follows:
\begin{align}
  \label{form:mocogan}
  \min_{G_{\rm I}} \max_{D_{\rm I}, D_{\rm V}}
  &\mathbb{E}_{{\bm x}}[\log{D_{\rm I}(x^{(t)})}]
  +\mathbb{E}_{\tilde{{\bm x}}}[\log{(1-D_{\rm I}(\tilde{x}^{(t)}))}] \nonumber \\
  &+\mathbb{E}_{{\bm x}}[\log{D_{\rm V}({\bm x})}]
  +\mathbb{E}_{\tilde{{\bm x}}}[\log{(1-D_{\rm V}(\tilde{{\bm x}}))}], 
\end{align}
where ${\bm x}$ and ${\tilde {\bm x}}$ represent the real video and the generated video, respectively.

\section{Proposed Model}
\label{sec:proposed}

In this study, we propose a conditional-MoCoGAN, which is a GAN architecture based on MoCoGAN, and it learns a conditional generative model that subdivides the latent space into classes by providing more detailed class information on motion and content.
An overview of the conditional-MoCoGAN architecture is given in Fig.\ref{fig:ourmodel}.
In the manner of the CGAN, we input the class label into the generator and the discriminator.
Through this approach, we believe that class-conditional video generation is possible.

\subsection{Generator}
The generator consists of a Gated Recurrent Unit (GRU) $R_{\rm M}$ and an image generator $G_{\rm I}$.
In conditional-MoCoGAN, the one-hot vector of the motion class label ${y}_{\rm m}$, a noise vector $\epsilon^{(t)}_{\rm m}$, and $z^{(t-1)}_{\rm m}$ are input into $R_{\rm M}$.
\begin{eqnarray}
  z^{(t)}_{\rm m} = R_{\rm M}(\epsilon^{(t)}_{\rm m} \oplus {y}_{\rm m}, z^{(t-1)}_{\rm m}),
\end{eqnarray}
where the operator $\oplus$ indicates the concatenation of vectors or tensors.
Additionally, the one-hot vector of the content class label ${y}_{\rm c}$, a noise vector $\epsilon_{\rm c}$, and $z^{(t)}_{\rm m}$ are input into $G_{\rm I}$.
\begin{eqnarray}
  {\tilde x}^{(t)} = G_{\rm I}(\epsilon_{\rm c} \oplus {y}_{\rm c} \oplus z^{(t)}_{\rm m})
\end{eqnarray}
$\epsilon^{(t)}_{\rm m},\epsilon_{\rm c}$ are sampled from Gaussian distribution. 
A generated video is equal to the sequence of frames and is represented by $\tilde{\bm x} = [{\tilde x}^{(1)}, {\tilde x}^{(2)}, \cdots, {\tilde x}^{(T)}].$

\subsection{Discriminator}
The whole discriminator consists of two discriminators, $D_{\rm I}$ and $D_{\rm V}$, similar to MoCoGAN.
Our model conditions the motion and content classes on a single video. Therefore, every video has a motion class label and a content class label.
Following the CGAN learning method, our discriminators receive the video and the class label concatenated as an input.
The inputs of the two discriminators are the sampled frame $x^{(t)}$ and the one-hot tensor of the content class label $y_{\rm c}$,
\begin{eqnarray}
  p_{\rm real} = D_{\rm I}(x^{(t)} \oplus y_{\rm c}) 
\end{eqnarray}
and the full video ${\bm x}$ and the one-hot tensor of the motion class $y_{\rm m}$,
\begin{eqnarray}
  p_{\rm real} = D_{\rm V}({\bm x} \oplus y_{\rm m}).
\end{eqnarray}

When an unknown combination of classes that are not included in the training data is specified, and a video of the corresponding class is correctly generated, we can say that zero-shot video generation has been achieved.

\section{Experiments}
\label{sec:exp}
\begin{figure}[h!]
  \begin{minipage}[b]{1.0\linewidth}
    \centering
    \centerline{\includegraphics[width=8.5cm]{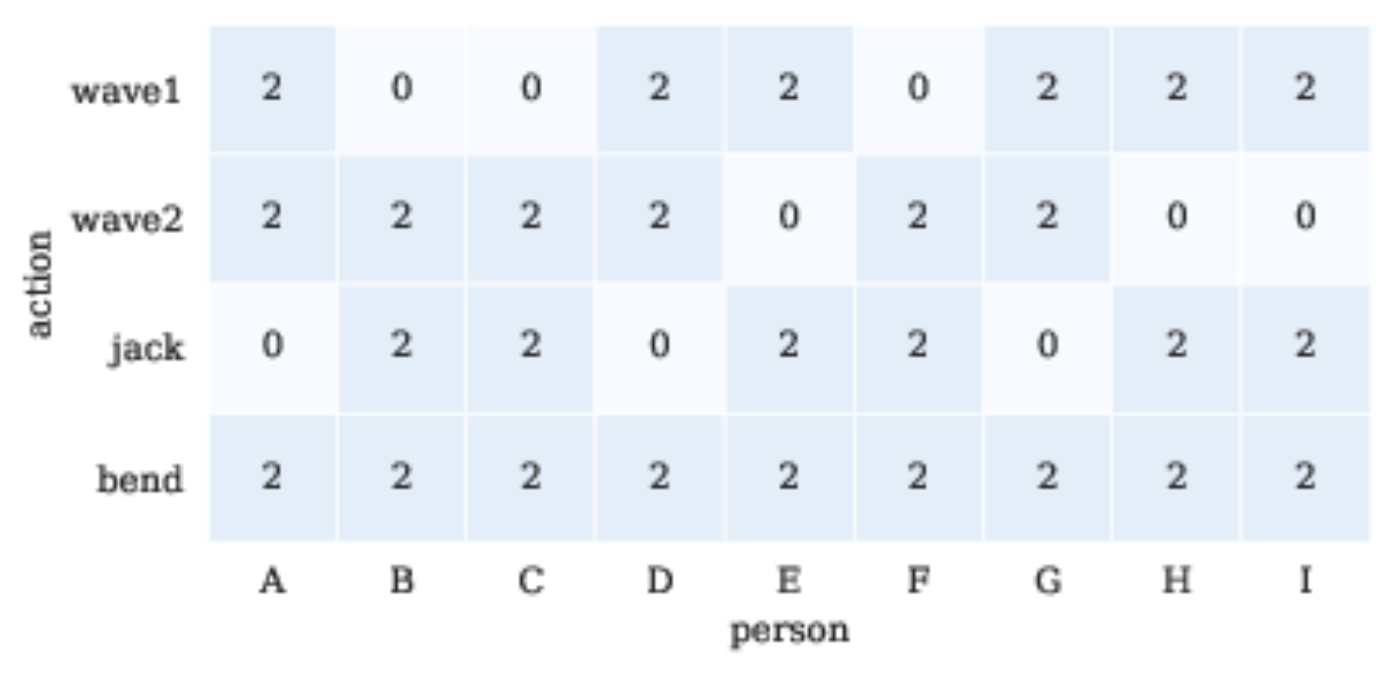}}
    \centerline{(a) real video}\medskip
  \end{minipage}
  \begin{minipage}[b]{1.0\linewidth}
    \centering
    \centerline{\includegraphics[width=8.5cm]{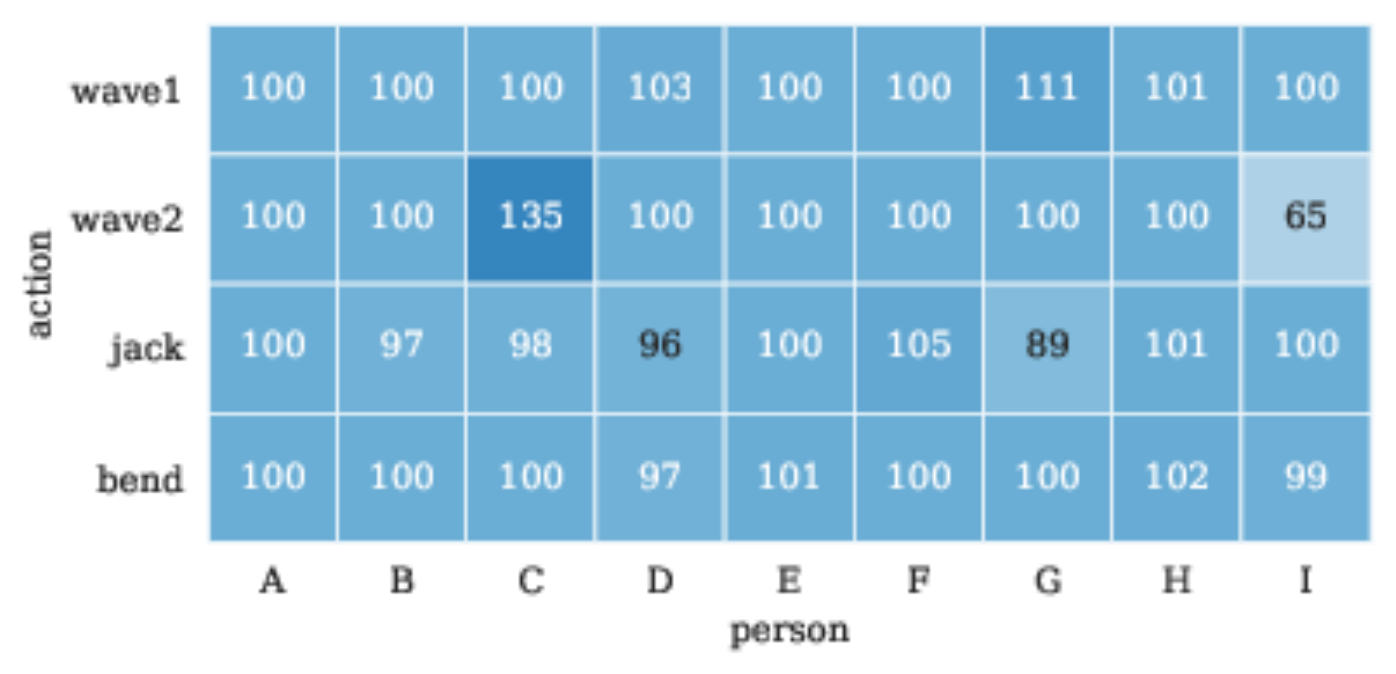}}
    \centerline{(b) ours}\medskip
  \end{minipage}
  \begin{minipage}[b]{1.0\linewidth}
    \centering
    \centerline{\includegraphics[width=8.5cm]{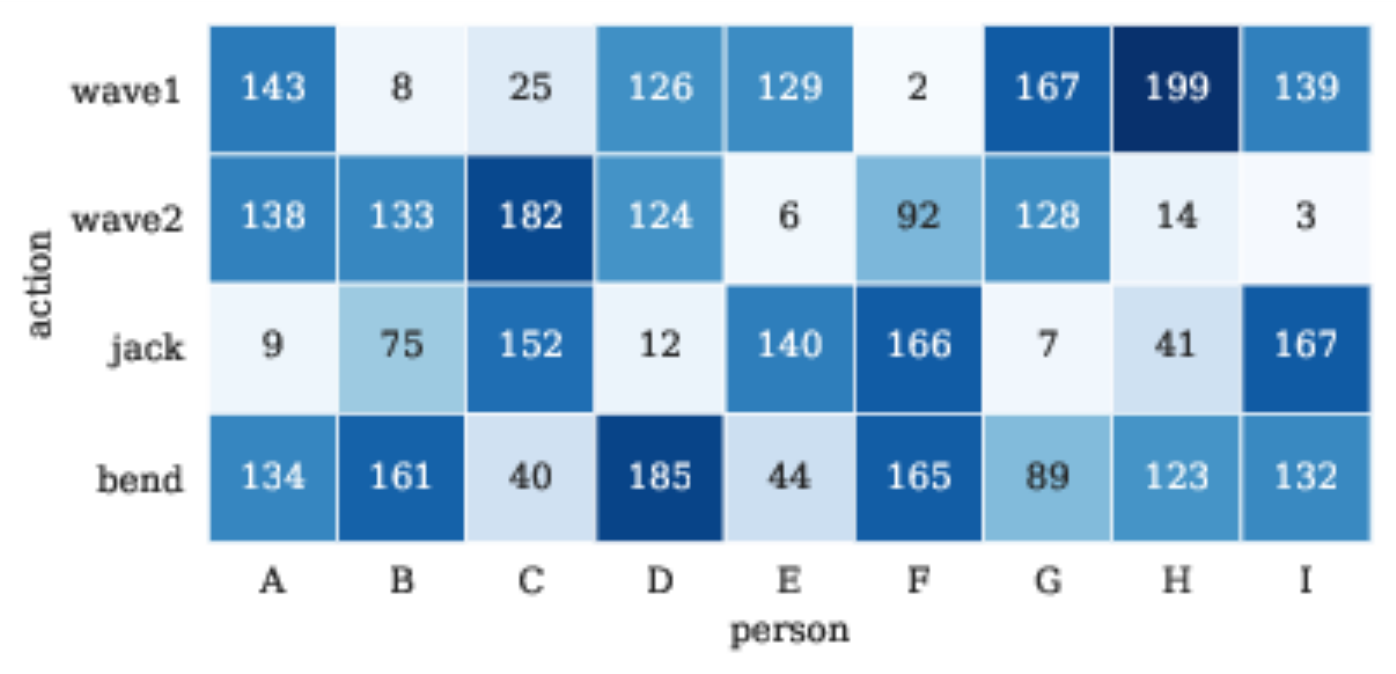}}
    \centerline{(c) MoCoGAN}\medskip
  \end{minipage}
  \caption{{Comparison of video classification results for $C_m$ and $C_c$ on the Weizmann action database.} A similar result was obtained for $C_d$. (a) indicates all classes included in the training data. (b)Results for 100 generated videos for all classes using our model. (c)Results for the same number of generated videos as in (b) using MoCoGAN.}
  \label{fig:res:weizmann:acc}
\end{figure}

\begin{table}[t]
  \begin{center}
    \caption{{Quantitative results}}
    \begin{tabular}{c|cccc} \hline
      \multicolumn{1}{c|}{} & \multicolumn{2}{c}{Weizmann} & \multicolumn{2}{c}{MUG} \\
      model & FID & Accuracy & FID & Accuracy\\ \hline \hline
      MoCoGAN & 60.04 & N/A & 30.95 & N/A \\
      ours & 103.29 & 98.7 & 28.83 & 96.4 \\ \hline
    \end{tabular}
    \label{tab:quantitative}
  \end{center}
\end{table}

We train the proposed model using the standard back propagation algorithm.
We use rectified linear unit (ReLU) as the activation function for $G_{\rm I}$ and leaky-ReLU for $D_{\rm I}$ and $D_{\rm V}$.
We employ the ADAM optimizer and set the learning rate to $2 \times 10^{-4}$ for all networks, and we apply spectral normalization \cite{spectral} to $D_{\rm I}$ and $D_{\rm V}$.
Dimensions of latent vectors $z_c$ and $z_m^{(t)}$ are both set to 30, following MoCoGAN \cite{mocogan}.
Our source code is available at GitHub\footnote{https://github.com/tj16kimura/Conditional-MoCoGAN}.

\subsection{Dataset}
We evaluate our model on the following datasets.
The length of videos is fixed at 16 frames.

The {Weizmann action database} \cite{weizmann} contains videos of nine different actions performed by nine people.
Motion classes include actions, such as jumping and waving-hands.
In this experiment, only the four action classes where the person does not move are used as training data.
We resized the videos to $64 \times 64$ pixels.

The {MUG facial expression database} \cite{mug} contains videos of seven different expressions performed by 86 people.
Motion classes include facial expressions, such as sad and surprised.
In this experiment, we use only the four action classes and nine content classes to maintain the same conditions as those in the experiment with the other dataset. 
We resized the videos to $96 \times 96$ pixels.

Additionally, one motion class from each content class is removed from the training data to evaluate zero-shot video generation.

\subsection{Quantitative Evaluation}
We used the classification accuracy of the generated videos by a classifier and the Fr\'{e}chet inception distance (FID) \cite{fid} as the quantitative evaluation metrics.

The FID is calculated as follows: 
\begin{eqnarray}
{||\mu - \tilde{\mu}||}^2 + {\rm Tr}(\Sigma + \tilde{\Sigma} - 2\sqrt{\Sigma \tilde{\Sigma}}),
\end{eqnarray}
where $\mu$ and $\Sigma$ represent the mean and covariance matrix computed from the real videos, and $\tilde{\mu}$ and $\tilde{\Sigma}$ represent the mean and covariance matrix computed from the generated videos.
We computed the FID between 100 generated videos and the videos in the dataset for all classes.

To calculate the classification accuracy, we trained the classifier separately on each dataset.
Next, we calculated the accuracy of motion and content class predicted by the classifier.
We trained three classifiers: $C_{\rm m}$ for only the motion class, $C_{\rm c}$ for only the content class, and $C_{\rm d}$ for both the motion and content classes.

The comparison results are shown in Table \ref{tab:quantitative} and Fig.\ref{fig:res:weizmann:acc}.
In Fig.\ref{fig:res:weizmann:acc}(a), classes with a value of 0 are the classes that are not included in the training data.
Although the FID does not improve, the balance of classes in the generated video is clearly better than that of the original MoCoGAN.

\begin{figure}[t]
  \begin{minipage}[b]{1.0\linewidth}
    \centering
    \centerline{\includegraphics[width=8.5cm]{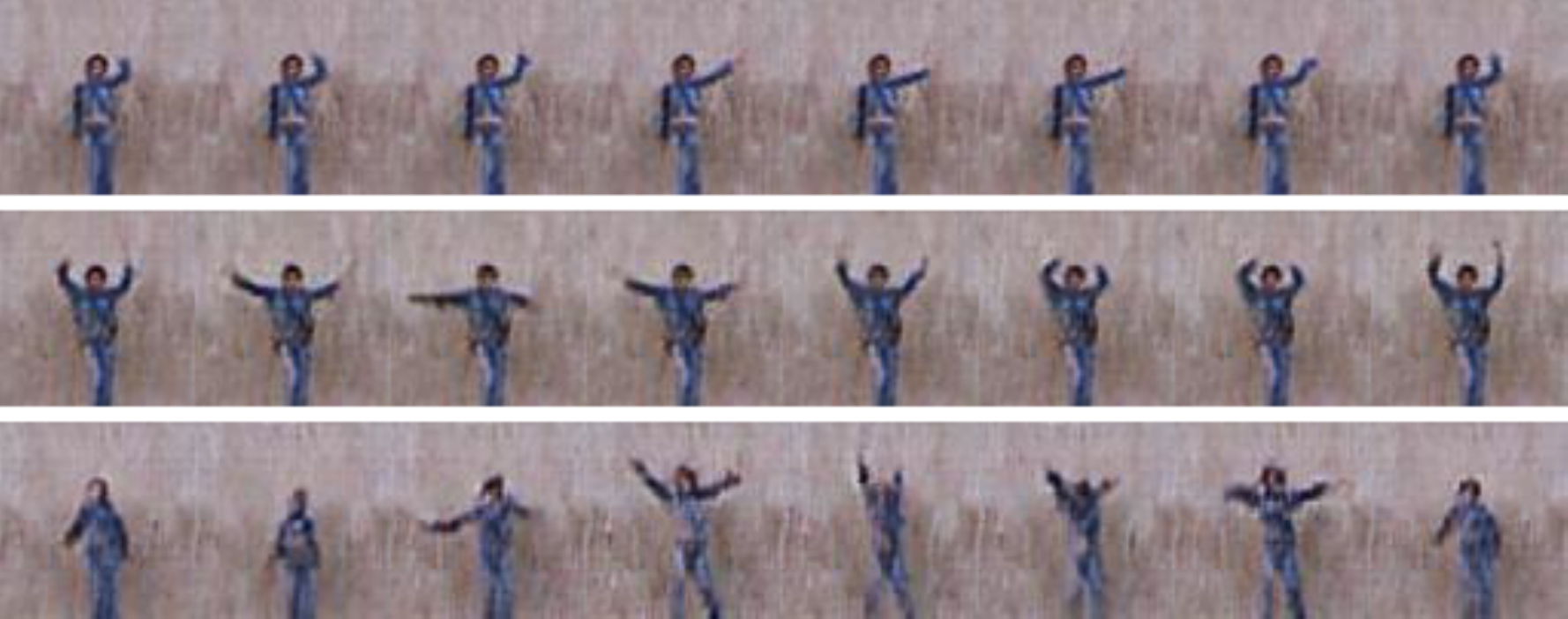}}
    \centerline{(a) ours}\medskip
  \end{minipage}
  \begin{minipage}[b]{1.0\linewidth}
    \centering
    \centerline{\includegraphics[width=8.5cm]{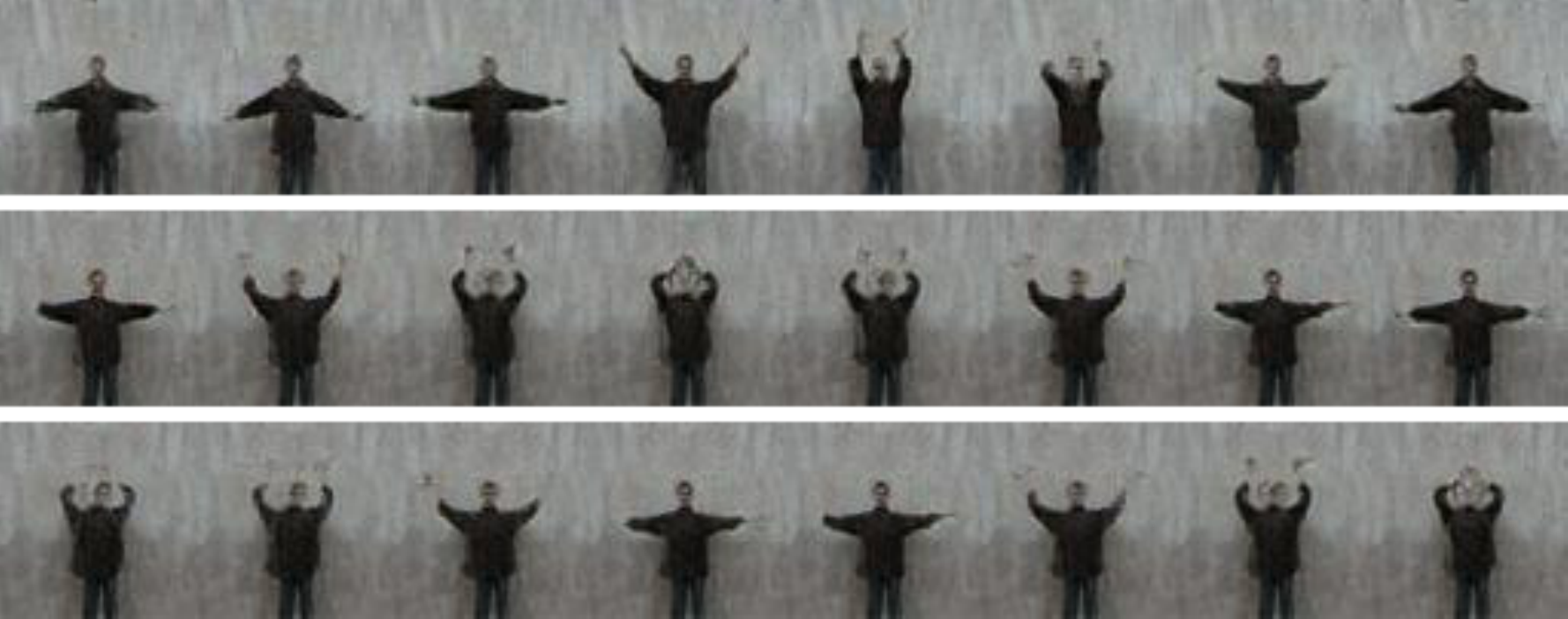}}
    \centerline{(b) MoCoGAN}\medskip
  \end{minipage}
  \caption{{Comparison of generated videos from fixed $z_c$.}}
  \label{fig:res:weizmann:zc}
\end{figure}

\begin{figure}[t]
  \begin{minipage}[b]{1.0\linewidth}
    \centering
    \centerline{\includegraphics[width=8.5cm]{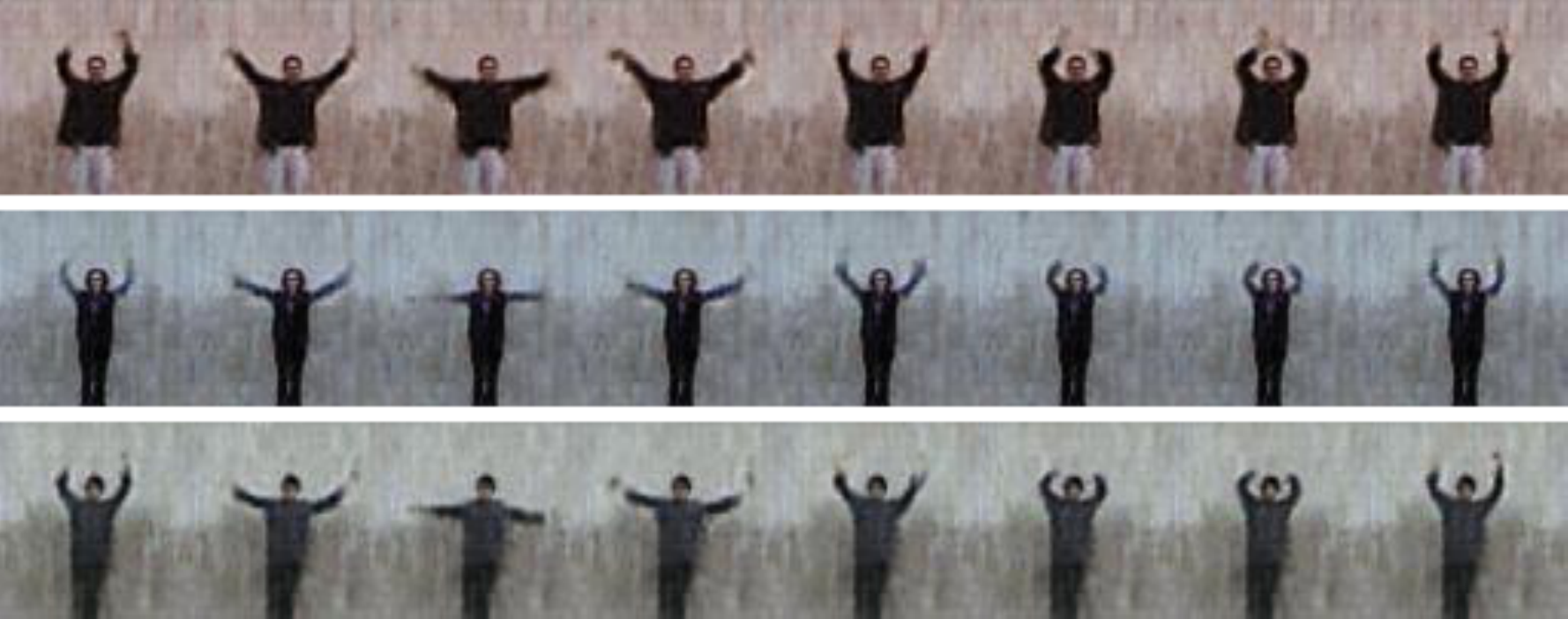}}
    \centerline{(a) ours}\medskip
  \end{minipage}
  \begin{minipage}[b]{1.0\linewidth}
    \centering
    \centerline{\includegraphics[width=8.5cm]{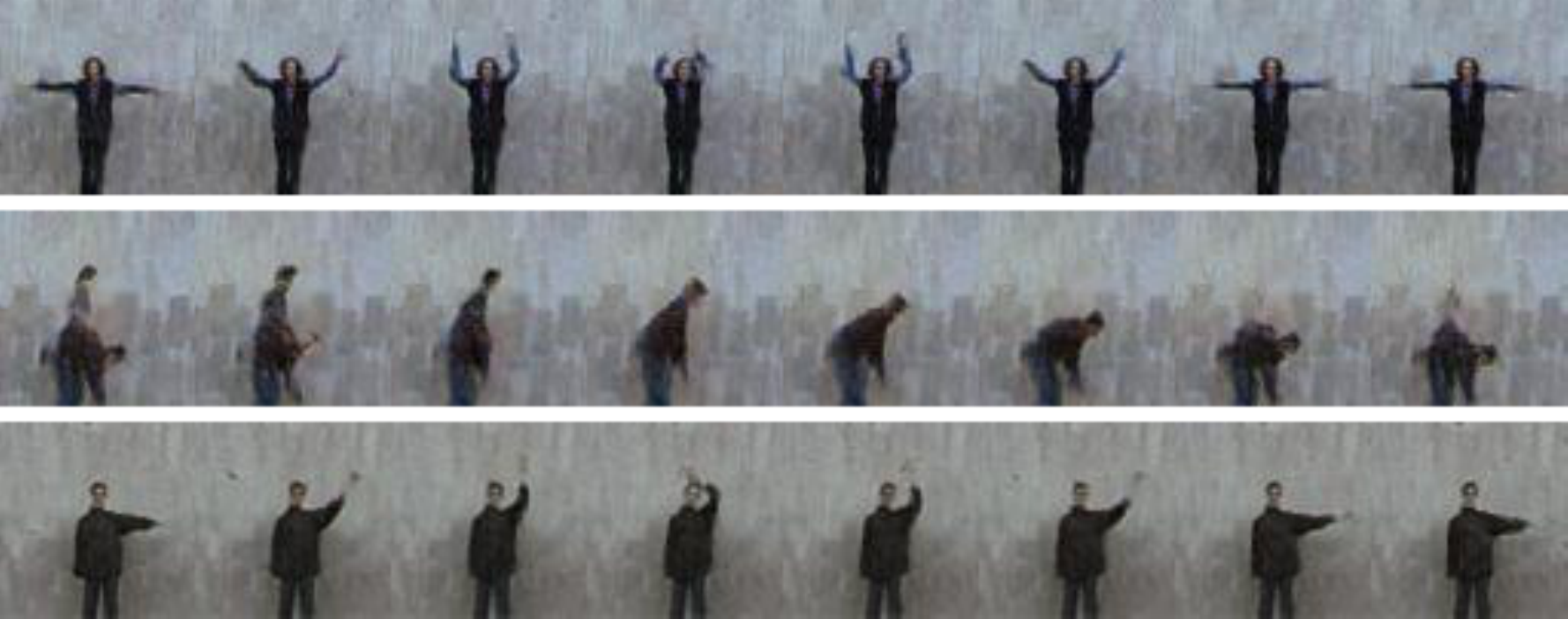}}
    \centerline{(b) MoCoGAN}\medskip
  \end{minipage}
  \caption{{Comparison of generated videos from fixed $z_m$.}}
  \label{fig:res:weizmann:zm}
\end{figure}
\subsection{Qualitative Evaluation}

In MoCoGAN, a video is generated from two vectors, $z_m$ and $z_c$.
As a qualitative evaluation, Fig.\ref{fig:res:weizmann:zc} and Fig.\ref{fig:res:weizmann:zm} show the generated videos when one of the vectors is fixed.

The latent space is separated into motion and content in MoCoGAN.
However, as shown in Fig.\ref{fig:res:weizmann:zm}(b), when $z_m$ was fixed and only $z_c$ was changed, both the motion and the content of the generated video changed.
On the contrary, when $z_c$ was fixed and only $z_m$ was changed, the motion of the generated video did not change.
This might be due to the fact that the separation of latent space is too simple, and in fact, the latent space has not been fully disentangled.
In our model, we solved this problem by further subdividing the latent space of motion and content into classes.
As shown in Fig.\ref{fig:res:weizmann:zc}(a) and Fig.\ref{fig:res:weizmann:zm}(a), when only one of the vectors, $z_c$ or $z_m$, is changed, the generated video changes correspondingly.
In particular, the video in the second row of Fig.\ref{fig:res:weizmann:zc}(a) corresponds to zero-shot video generation.

\section{Conclusion}
\label{sec:conclusion}
In this study, we propose a conditional video generative model that subdivides the latent space into classes by incorporating the learning method of CGAN into MoCoGAN. The MoCoGAN is a video generative model that divides the latent space into motion and content.
We trained our model on two datasets and evaluated the generated videos using quantitative and qualitative methods.
From the results, we found that class-conditional generation is possible in video generation, and that zero-shot generation, which generates a class with a combination of multiple class conditionals that do not exist in the training data, is also possible. 
Alternatively, the quality of the generated video decreases and needs to be improved.

%

\bibliographystyle{IEEEbib}
\bibliography{strings,refs}

\begin{thebibliography}{10}

\bibitem{gan}
Ian Goodfellow, Jean Pouget-Abadie, Mehdi Mirza, Bing Xu, David Warde-Farley,
  Sherjil Ozair, Aaron Courville, and Yoshua Bengio,
\newblock ``Generative adversarial nets,''
\newblock {\em in NeurIPS}, vol. 27, pp. 2672--2680, 2014.

\bibitem{videogan}
Carl Vondrick, Hamed Pirsiavash, and Antonio Torralba,
\newblock ``Generating videos with scene dynamics,''
\newblock in {\em NeurIPS}, 2016, pp. 613--621.

\bibitem{mocogan}
Sergey Tulyakov, Ming-Yu Liu, Xiaodong Yang, and Jan Kautz,
\newblock ``Mocogan: Decomposing motion and content for video generation,''
\newblock in {\em CVPR}, 2018, pp. 1526--1535.

\bibitem{distillgan}
KJ~Joseph, Arghya Pal, Sailaja Rajanala, and Vineeth~N Balasubramanian,
\newblock ``Zero-shot image generation by distilling concepts from multiple
  captions,''
\newblock in {\em ICML Workshop}, 2018.

\bibitem{hoigan}
Megha Nawhal, Mengyao Zhai, Andreas Lehrmann, Leonid Sigal, and Greg Mori,
\newblock ``Generating videos of zero-shot compositions of actions and
  objects,''
\newblock in {\em ECCV}. Springer, 2020, pp. 382--401.

\bibitem{ebdn}
Caroline Chan, Shiry Ginosar, Tinghui Zhou, and Alexei~A Efros,
\newblock ``Everybody dance now,''
\newblock in {\em ICCV}, 2019, pp. 5933--5942.

\bibitem{cgan}
Mehdi Mirza and Simon Osindero,
\newblock ``Conditional generative adversarial nets,''
\newblock {\em arXiv preprint arXiv:1411.1784}, 2014.

\bibitem{spectral}
Takeru Miyato, Toshiki Kataoka, Masanori Koyama, and Yuichi Yoshida,
\newblock ``Spectral normalization for generative adversarial networks,''
\newblock in {\em ICLR}, 2018.

\bibitem{weizmann}
Moshe Blank, Lena Gorelick, Eli Shechtman, Michal Irani, and Ronen Basri,
\newblock ``Actions as space-time shapes,''
\newblock in {\em ICCV}, 2005, vol.~2, pp. 1395--1402.

\bibitem{mug}
Niki Aifanti, Christos Papachristou, and Anastasios Delopoulos,
\newblock ``The mug facial expression database,''
\newblock in {\em WIAMIS10}, 2010, pp. 1--4.

\bibitem{fid}
Martin Heusel, Hubert Ramsauer, Thomas Unterthiner, Bernhard Nessler, and Sepp
  Hochreiter,
\newblock ``Gans trained by a two time-scale update rule converge to a local
  nash equilibrium,''
\newblock in {\em NeurIPS}, 2017, pp. 6626--6637.

\end{thebibliography}

\end{document}